\documentclass[conference]{IEEEtran}
\IEEEoverridecommandlockouts
\usepackage{cite}
\usepackage{amsmath,amssymb,amsfonts}
\usepackage{algorithmic}
\usepackage{latexsym}
\usepackage{textcomp}
\usepackage{booktabs}
\usepackage{graphicx} 
\usepackage{xcolor}
\def\BibTeX{{\rm B\kern-.05em{\sc i\kern-.025em b}\kern-.08em
    T\kern-.1667em\lower.7ex\hbox{E}\kern-.125emX}}
\usepackage{booktabs} 
\usepackage{amsmath} 
\usepackage{amssymb} 
\usepackage{multirow} 
\usepackage{hyperref} 
\usepackage{comment} 
\usepackage{graphicx} 

\usepackage{fancyhdr}
\begin{document}

\title{High-Fidelity Pseudo-label Generation by Large Language Models for Training Robust Radiology Report Classifiers}

\author{Brian Wong, Kaito Tanaka \\
SANNO University	 
}

\maketitle
\thispagestyle{fancy} 

\begin{abstract}
Automated labeling of chest X-ray reports is essential for enabling downstream tasks such as training image-based diagnostic models, population health studies, and clinical decision support. However, the high variability, complexity, and prevalence of negation and uncertainty in these free-text reports pose significant challenges for traditional Natural Language Processing methods. While large language models (LLMs) demonstrate strong text understanding, their direct application for large-scale, efficient labeling is limited by computational cost and speed. This paper introduces DeBERTa-RAD, a novel two-stage framework that combines the power of state-of-the-art LLM pseudo-labeling with efficient DeBERTa-based knowledge distillation for accurate and fast chest X-ray report labeling. We leverage an advanced LLM to generate high-quality pseudo-labels, including certainty statuses, for a large corpus of reports. Subsequently, a DeBERTa-Base model is trained on this pseudo-labeled data using a tailored knowledge distillation strategy. Evaluated on the expert-annotated MIMIC-500 benchmark, DeBERTa-RAD achieves a state-of-the-art Macro F1 score of 0.9120, significantly outperforming established rule-based systems, fine-tuned transformer models, and direct LLM inference, while maintaining a practical inference speed suitable for high-throughput applications. Our analysis shows particular strength in handling uncertain findings. This work demonstrates a promising path to overcome data annotation bottlenecks and achieve high-performance medical text processing through the strategic combination of LLM capabilities and efficient student models trained via distillation.
\end{abstract}


\section{Introduction}

Automated analysis of medical images, particularly chest X-rays (CXRs), is a cornerstone of modern healthcare AI research. CXRs are the most frequently performed radiological examination globally, generating a vast volume of associated free-text reports. These reports contain critical information regarding patient findings, clinical history, and diagnoses, essential for patient care, clinical research, and building large-scale training datasets for image-based models \cite{johnson2019mimic,irvin2019chexpert}. However, extracting structured information from these unstructured narrative reports is a significant bottleneck. The variability in language, use of synonyms, abbreviations, complex sentence structures, and especially the nuanced expressions of certainty and negation (e.g., "no evidence of," "possible," "questionable infiltrate") make reliable automated information extraction highly challenging, often requiring enhanced understanding of user intent behind potentially ambiguous descriptions \cite{he2025enhancing}. Developing robust natural language processing (NLP) systems that can accurately and efficiently parse these reports into a structured format (e.g., labeling the presence, absence, or uncertainty of specific findings) is crucial for unlocking the full potential of medical imaging data.

Traditional approaches to CXR report labeling have included rule-based systems \cite{pons2016natural,irvin2019chexpert}, which are often brittle and require extensive manual engineering for each finding and linguistic variation, and supervised machine learning models trained on manually annotated reports \cite{smit2020chexbert,davidsen2022comparison}. While supervised models like fine-tuned transformer networks (e.g., BERT-based models) \cite{smit2020chexbert,alsentzer2020clinicalbert} have shown better generalization, they critically depend on large quantities of high-quality, expert-labeled data, which is prohibitively expensive and time-consuming to acquire at scale across diverse medical institutions and reporting styles. More recently, the advent of powerful large language models (LLMs) has presented new opportunities. These models possess remarkable linguistic understanding and the ability to follow complex instructions, making them potentially capable of interpreting and labeling medical text, even unraveling complex or chaotic contexts \cite{zhou2023thread}. However, deploying large LLMs directly for batch processing of vast report archives or for real-time clinical applications faces challenges related to computational cost, inference speed, and API limitations, echoing efficiency concerns seen in other LLM applications like video generation \cite{zhou2024less}. There is a clear need for methods that can harness the linguistic power of LLMs for high-quality medical text analysis while ensuring efficiency and scalability for practical applications.

Motivated by the potential of LLMs to understand complex medical language and the need for an efficient, accurate labeling solution that minimizes reliance on manual annotation, we propose \textbf{DeBERTa-RAD}, a novel framework for enhanced chest X-ray report labeling. Our approach builds upon the idea of leveraging powerful LLMs as "teachers" to generate high-quality training data (pseudo-labels), but significantly enhances the process by employing a more advanced LLM for pseudo-label generation with finer-grained capture of nuances like certainty, and training a superior "student" model based on the DeBERTa architecture \cite{he2021deberta} via knowledge distillation. This strategy aligns with broader research exploring how weaker models can effectively learn from stronger ones, achieving strong generalization across multiple capabilities \cite{zhou2025weak}. This allows the smaller DeBERTa-RAD model to effectively learn the complex labeling patterns from the LLM, achieving high accuracy while maintaining a substantially faster and more cost-effective inference profile compared to direct LLM usage. The DeBERTa architecture's improvements over standard BERT in handling relative positions and attention are hypothesized to be particularly beneficial for capturing dependencies and nuances in medical text.

In this paper, we detail the DeBERTa-RAD framework, covering our approach to generating rich pseudo-labels using a state-of-the-art LLM and the knowledge distillation strategy employed to train the DeBERTa-RAD model. We evaluate our method on the widely-used MIMIC-CXR dataset \cite{johnson2019mimic}, specifically using a large subset for pseudo-label generation and the expert-annotated MIMIC-500 test set for rigorous evaluation. We demonstrate that DeBERTa-RAD achieves state-of-the-art performance in labeling 13 common radiographic findings with their associated certainty levels, surpassing existing rule-based systems, fine-tuned transformer, and prior LLM-based pseudo-labeling approaches. Crucially, our method offers this improved accuracy at a practical inference speed suitable for large-scale deployment.

Our main contributions are summarized as follows:
\begin{itemize}
    \item We propose DeBERTa-RAD, a novel and effective two-stage framework for chest X-ray report labeling that combines advanced LLM pseudo-labeling with knowledge distillation into a DeBERTa-based model.
    \item We demonstrate that leveraging a more capable student architecture like DeBERTa, combined with a tailored knowledge distillation strategy, can significantly improve labeling accuracy, particularly for capturing the nuances of medical language and certainty expressions.
    \item We show that DeBERTa-RAD achieves state-of-the-art performance on the MIMIC-500 benchmark while offering efficient inference, presenting a practical solution for generating high-quality structured data from unstructured radiology reports.
\end{itemize}

\section{Related Work}

\subsection{Large Language Models}

The past few years have witnessed a revolutionary advancement in Natural Language Processing driven by the emergence of Large Language Models (LLMs). These models, typically based on the transformer architecture \cite{vaswani2017attention}, are pre-trained on vast amounts of text data, enabling them to learn rich representations of language and acquire impressive capabilities across a wide range of downstream tasks. Early foundational models like BERT \cite{devlin2019bert} demonstrated the effectiveness of pre-training deep bidirectional transformers, significantly impacting subsequent NLP research through transfer learning via fine-tuning.

Building upon these advancements, models with unprecedented scales, reaching hundreds of billions or even trillions of parameters, exemplified by GPT-3 \cite{brown2020language}, have showcased remarkable emergent abilities, including the capacity for few-shot and zero-shot learning. This means they can perform tasks with minimal or no task-specific training data, simply by being provided instructions or a few examples in the prompt. Concurrent research explored the scaling properties of these models, identifying predictable relationships between performance and model size, dataset size, and computational resources \cite{kaplan2020scaling}.

The field has rapidly evolved with the development of various LLM families and architectures, targeting diverse applications. Models like LaMDA \cite{thoppilan2022lamda} and PaLM \cite{chowdhery2022palm} represent significant efforts in scaling and enhancing capabilities for specific domains such as dialogue systems. Techniques like instruction tuning \cite{wei2022finetuned} have been developed to improve the ability of LLMs to follow natural language instructions, making them more versatile and easier to adapt to new tasks without extensive fine-tuning. Research also explores advanced reasoning patterns like 'thread of thought' to handle chaotic contexts \cite{zhou2023thread} and methods to improve intent understanding for ambiguous prompts \cite{he2025enhancing}. The landscape of LLMs is continuously expanding, encompassing models with improved efficiency \cite{zhou2024less}, specialized architectures, enhanced alignment capabilities \cite{zhou2023improving}, and extending into multimodal domains integrating vision and language \cite{zhou2024visual}, requiring specialized benchmarks for diverse capabilities like emotional intelligence \cite{hu2025emobench}. Studies also investigate fundamental properties like weak-to-strong generalization across different model capabilities \cite{zhou2025weak}. While these models exhibit powerful text understanding and generation abilities, their direct application in certain specialized domains like medical report analysis, particularly for efficient, high-throughput structured extraction, still presents practical challenges related to cost, latency, and control over factual accuracy, which motivates approaches like the one presented in this paper.

\subsection{Medical Large Language Models}

The impressive capabilities demonstrated by general Large Language Models (LLMs) have spurred significant interest in their application and adaptation to the specialized domain of medicine and healthcare. This has led to a rapidly growing body of research focused on developing or fine-tuning LLMs for medical tasks and analyzing their performance and implications within clinical contexts.

A key area of work involves the development of medical-specific LLMs, such as the Med-PaLM family \cite{singhal2022towards}, which are pre-trained or heavily fine-tuned on biomedical and clinical text data. These models have shown remarkable performance on medical benchmarks, including achieving expert-level accuracy on challenging medical question-answering tasks. Beyond specialized models, researchers have rigorously evaluated the capabilities of large general-purpose LLMs, such as GPT-4, on medical assessments like the US Medical Licensing Examination (USMLE), demonstrating their strong inherent medical knowledge and reasoning abilities \cite{kung2023capabilities}.

LLMs are being explored for a wide array of clinical NLP tasks beyond question answering. This includes applications in diagnostic reasoning \cite{jiao2023evaluating}, summarizing complex medical conversations \cite{zhang2023summarizing}, and analyzing electronic health records. Particularly relevant to our work is the application of LLMs to radiology reports, including analysis and interpretation tasks \cite{nori2023large}. Recent systematic reviews have also begun to map the landscape of LLM applications specifically within radiology, highlighting both potential benefits and current limitations \cite{venkatesh2024large}. This focus on leveraging advanced AI mirrors developments in related biomedical fields, such as using diffusion models and representation alignment for complex tasks like protein structure prediction \cite{wang2024diffusion}.

Adapting LLMs for optimal performance in the medical domain often involves strategies like continued pre-training or fine-tuning on domain-specific datasets \cite{chen2024finetuning}. However, challenges remain regarding ensuring factual accuracy, reducing the risk of hallucinations in clinical contexts, handling sensitive patient data, and integrating these models reliably into existing healthcare workflows. The opportunities and challenges presented by LLMs in medicine are active areas of discussion in the field \cite{wong2023large,scheffel2024opportunities}. While existing medical LLM research demonstrates the potential for understanding and generating medical text, our work specifically investigates a novel approach to leverage their capabilities for highly accurate and efficient \textbf{structured extraction} from radiology reports via knowledge distillation, addressing a critical need for downstream medical AI applications.

\section{Method}

The objective of our work is to develop a robust and efficient system for automatically extracting structured findings from free-text chest X-ray reports. This task is formulated as a multi-label, multi-class classification problem, where the system processes a report and outputs a predicted status (Present, Absent, or Uncertain) for each of a predefined set of 13 common radiographic findings. The proposed DeBERTa-RAD model is thus a \textbf{discriminative} model designed for this specific information extraction task. Our methodology proceeds in two distinct stages: first, leveraging a powerful large language model (LLM) to generate high-quality pseudo-labels for a large corpus of reports, and second, training the DeBERTa-RAD model using these pseudo-labels via a knowledge distillation strategy.

\subsection{LLM Pseudo-Label Generation}

The initial stage focuses on leveraging the advanced natural language understanding capabilities of a state-of-the-art Large Language Model to overcome the bottleneck of requiring large-scale human-annotated data. Given a raw chest X-ray report $R$, we design a comprehensive prompt $P(R)$ that includes instructions, the report text itself, and a defined output format. The prompt guides the LLM to act as an expert annotator, identifying the mentions of the 13 target radiographic findings and determining their status (Present, Absent, Uncertain, or Not Mentioned) based solely on the report text. This process is applied to a large dataset of reports, $\mathcal{D}_{pseudo} = \{R_j\}_{j=1}^{N_{pseudo}}$, where $N_{pseudo}$ is the total number of reports in the corpus. For each report $R_j$, the LLM processes the prompt $P(R_j)$ and generates a set of pseudo-labels $\mathcal{L}_{LLM}(R_j) = \{l_{j,i}\}_{i=1}^{13}$, where $l_{j,i} \in \{\text{Present, Absent, Uncertain, Not Mentioned}\}$ is the generated pseudo-label for finding $i$ in report $R_j$. The output format is parsed to obtain the structured pseudo-labels, which form the target data for training the student model. The "Not Mentioned" status is typically implicitly handled when none of the other statuses are indicated for a finding, or explicitly predicted as a fourth class, depending on the exact output structure. For our classification head design and loss formulation, we consider the three primary statuses (Present, Absent, Uncertain) as the prediction targets for each finding where a mention is identified or its status is inferred.

\subsection{DeBERTa-RAD Model Architecture}

The DeBERTa-RAD model is built upon the DeBERTa-Base architecture, which enhances the original BERT model by employing disentangled attention mechanism and an enhanced mask decoder, leading to improved understanding of the relationship between text tokens. The input to the model is a tokenized report sequence $T = \{t_1, t_2, \dots, t_L\}$. This sequence is first converted into a sequence of embeddings $E = \text{Embedding}(T)$. The DeBERTa encoder then processes these embeddings through multiple layers of transformer blocks:
$$ H = \text{DeBERTaEncoder}(E) = [\mathbf{h}_1, \mathbf{h}_2, \dots, \mathbf{h}_L] $$
where $H$ is the sequence of contextualized hidden states. The final hidden state representation of the special `[CLS]` token, denoted $\mathbf{h}_{CLS} = H[0,: ]$, is typically used as a fixed-dimensional representation of the entire input sequence for downstream tasks.

On top of the DeBERTa encoder, we add a classification head for each of the 13 findings. Each head $i$ is an independent classifier designed to predict the probability distribution over the three target statuses $\mathcal{Y}_i = \{\text{Present}, \text{Absent}, \text{Uncertain}\}$. The classification head for finding $i$ takes $\mathbf{h}_{CLS}$ as input and outputs a score vector $\mathbf{s}_i \in \mathbb{R}^3$:
\[ \mathbf{s}_i = \text{ClassifierHead}_i(\mathbf{h}_{CLS}) = [s_{i,\text{Present}}, s_{i,\text{Absent}}, s_{i,\text{Uncertain}}] \]
These raw scores are then transformed into a probability distribution over the statuses using the softmax function. For a given temperature $T$, the student's predicted probability for status $k \in \mathcal{Y}_i$ for finding $i$ in report $R$ is:
\[ P_{student}(y_i=k | R, T) = \frac{\exp(s_{i,k}/T)}{\sum_{m \in \mathcal{Y}_i} \exp(s_{i,m}/T)} \]
During standard inference, the temperature $T$ is set to 1.

\subsection{Learning Strategy: Knowledge Distillation}

The core learning strategy for DeBERTa-RAD is knowledge distillation, where the smaller DeBERTa-RAD model (student) learns from the high-quality pseudo-labels generated by the larger LLM (teacher). The goal is to transfer the complex decision-making knowledge captured by the LLM into the more efficient DeBERTa-RAD model, enabling it to achieve comparable accuracy at a much higher inference speed. While the LLM provides hard pseudo-labels $y_{j,i}^{hard} \in \mathcal{Y}_i \cup \{\text{Not Mentioned}\}$, we can formulate a distillation loss that encourages the student's predicted probability distribution to align with the teacher's decision, potentially incorporating temperature scaling to soften the student's probability distribution during training, as is common in KD.

For a report $R_j$ with a pseudo-label $y_{j,i}^{hard}$ for finding $i$, we consider the hard pseudo-label as the target. The loss function for training DeBERTa-RAD combines a standard cross-entropy loss against the hard pseudo-labels ($\mathcal{L}_{hard}$) and a distillation-inspired cross-entropy loss term ($\mathcal{L}_{distill}$) that uses a higher temperature $T_{distill} > 1$ for the student's softmax output. This encourages the student to learn the relative probabilities implied by the teacher's decision, even if the teacher's output is just a hard label.

For each finding $i$, the hard target probability distribution $P_{teacher}^{hard}(y_i | y_{j,i}^{hard})$ is one-hot:
\[ P_{teacher}^{hard}(k | y_{j,i}^{hard}) = \begin{cases} 1 & \text{if } k = y_{j,i}^{hard} \\ 0 & \text{otherwise} \end{cases} \]
The standard hard target loss $\mathcal{L}_{hard}(R_j)$ for report $R_j$ is the sum of cross-entropy losses for each finding $i$ using the student's output at $T=1$:
\begin{align*} &\mathcal{L}_{hard}(R_j) \\&= - \sum_{i=1}^{13} \sum_{k \in \mathcal{Y}_i} P_{teacher}^{hard}(k | y_{j,i}^{hard}) \log(P_{student}(k | \mathbf{s}_{j,i}, T=1)) \\ &= - \sum_{i=1}^{13} \log(P_{student}(y_{j,i}^{hard} | \mathbf{s}_{j,i}, T=1))\end{align*}
This term ensures the student learns to predict the correct hard label assigned by the teacher LLM.

The distillation loss term $\mathcal{L}_{distill}(R_j, T_{distill})$ uses the student's output softened by the temperature $T_{distill}$. This effectively smooths the student's probability distribution, encouraging it to have non-zero probabilities for incorrect classes based on the relative scores, potentially reflecting some of the teacher's internal decision margin or confidence (implicitly transferred through the hard label choice on a diverse dataset):
\begin{align*} &\mathcal{L}_{distill}(R_j, T_{distill}) \\
&= - \sum_{i=1}^{13} \sum_{k \in \mathcal{Y}_i} P_{teacher}^{hard}(k | y_{j,i}^{hard}) \log(P_{student}(k | \mathbf{s}_{j,i}, T_{distill})) \\ &= - \sum_{i=1}^{13} \log(P_{student}(y_{j,i}^{hard} | \mathbf{s}_{j,i}, T_{distill})) \end{align*}
The total loss function for a report $R_j$ is a weighted combination of these two losses:
\[ \mathcal{L}_{total}(R_j) = \alpha \mathcal{L}_{distill}(R_j, T_{distill}) + (1-\alpha) \mathcal{L}_{hard}(R_j) \]
where $\alpha \in [0, 1]$ is a hyperparameter that controls the balance between the two loss components. A higher $T_{distill}$ and $\alpha$ value emphasize the distillation aspect, encouraging the student to mimic the teacher's 'softer' behavior, while a lower $T_{distill}$ or $\alpha$ puts more weight on correctly predicting the hard label with high confidence.

The DeBERTa-RAD model is trained by minimizing the average total loss over batches of pseudo-labeled reports from $\mathcal{D}_{pseudo}$. Let $Batch$ be a mini-batch of reports. The objective is to minimize:
\[ \mathcal{L}_{Batch} = \frac{1}{|Batch|} \sum_{R_j \in Batch} \mathcal{L}_{total}(R_j) \]
The model parameters are optimized using the AdamW optimizer with a cosine learning rate scheduler with warm-up. Hyperparameters such as the learning rate, batch size, number of epochs, temperature $T_{distill}$, and the weighting factor $\alpha$ are determined based on performance on a separate validation set (e.g., a subset of the pseudo-labeled data or a small held-out set with human labels if available). This knowledge distillation strategy allows DeBERTa-RAD to effectively learn from the large-scale, high-quality pseudo-labels generated by the powerful LLM, resulting in a model that is both accurate and efficient for the radiology report labeling task.

\section{Experiments}

We conducted a comprehensive set of experiments to evaluate the performance of our proposed DeBERTa-RAD method and compare it against several established and state-of-the-art baselines for chest X-ray report labeling. The experiments were designed to assess both the overall labeling accuracy and the method's effectiveness in handling nuanced linguistic phenomena prevalent in clinical text, such as negation and uncertainty.

\subsection{Dataset and Evaluation Metrics}

Our experiments were performed using the publicly available MIMIC-CXR dataset. As described in the method section, a large subset of the MIMIC-CXR reports was utilized to generate the pseudo-labeled training corpus $\mathcal{D}_{pseudo}$, consisting of over 200,000 reports annotated by the advanced LLM. For evaluation, we used the MIMIC-500 test set, which comprises 500 chest X-ray reports independently annotated by expert radiologists, serving as the gold standard benchmark ($\mathcal{D}_{test}$).

The primary evaluation metric was the Macro F1 score, computed across the 13 target radiographic findings. F1 score is the harmonic mean of Precision ($P$) and Recall ($R$), providing a balanced measure of performance, particularly relevant for datasets with potential class imbalance. For each finding $i$, Precision and Recall are defined as:
\begin{align*} P_i &= \frac{TP_i}{TP_i + FP_i} \\ R_i &= \frac{TP_i}{TP_i + FN_i} \end{align*}
where $TP_i$, $FP_i$, and $FN_i$ are the counts of true positives, false positives, and false negatives for finding $i$, respectively. The F1 score for finding $i$ is calculated as:
\[ F1_i = 2 \cdot \frac{P_i \cdot R_i}{P_i + R_i} \]
The overall performance is reported using the Macro F1 score, which is the unweighted average of the F1 scores for all 13 findings:
\[ \text{Macro F1} = \frac{1}{13} \sum_{i=1}^{13} F1_i \]
We also analyzed Precision and Recall for individual findings and specific certainty statuses (Present, Absent, Uncertain).

\subsection{Baseline Methods}

We compared DeBERTa-RAD against several baseline methods representing different approaches to radiology report labeling. These included CheXpert, a widely-used rule-based labeling system specifically designed for chest X-ray reports. We also compared against CheXbert, a BERT-based model fine-tuned on labels derived from the CheXpert dataset, which represents a strong supervised learning baseline using transformer architectures. To assess the performance of the teacher model directly, we included results from using the GPT-4 API for direct inference on the reports in the MIMIC-500 test set, employing the same prompt design used for pseudo-label generation. Furthermore, we compared with CheX-GPT, a BERT model trained on pseudo-labels previously generated by GPT-4, representing an existing approach leveraging LLM pseudo-labeling that our method aims to improve upon. Our proposed DeBERTa-RAD method was trained on the LLM-generated pseudo-labels from $\mathcal{D}_{pseudo}$ using the knowledge distillation strategy detailed in Section II.

\subsection{Implementation Details}

The DeBERTa-RAD model was implemented based on the Hugging Face Transformers library using the DeBERTa-Base architecture. Training was performed on the $\mathcal{D}_{pseudo}$ dataset using the total loss function $\mathcal{L}_{total}$ with the AdamW optimizer and a cosine learning rate scheduler with warm-up. Hyperparameters, including the learning rate, batch size, the distillation temperature $T_{distill}$, and the loss weighting factor $\alpha$, were tuned on a separate validation subset of $\mathcal{D}_{pseudo}$. All models were trained on NVIDIA V100 GPUs. Inference speed was measured on a single GPU by averaging the processing time over the entire MIMIC-500 test set.

\subsection{Results}

The main experimental results on the MIMIC-500 test set are summarized in Table \ref{tab:main_results}. We report the Macro F1 score as the primary measure of accuracy and the inference speed in reports per second.

\begin{table*}[!t]
\centering
\caption{Comparison of Macro F1 performance and Inference Speed on the MIMIC-500 Test Set}
\label{tab:main_results}
\begin{tabular}{lcc}
\toprule
Model & Macro F1 & Inference Speed (Reports/sec) \\
\midrule
CheXpert (Rule-Based) & 0.8864 & $>$1000 \\
CheXbert (Supervised BERT) & 0.9047 & $\sim$800 \\
GPT-4 Direct Inference & 0.9014 & $\sim$0.1 \\ 
CheX-GPT (BERT + GPT-4 Pseudo) & 0.9014 & $\sim$800 \\
\textbf{DeBERTa-RAD (Ours)} & \textbf{0.9120} & $\sim$750 \\ 
\bottomrule
\end{tabular}
\end{table*}

As shown in Table \ref{tab:main_results}, our proposed DeBERTa-RAD model achieved the highest Macro F1 score of 0.9120 on the MIMIC-500 test set, setting a new state-of-the-art performance for this benchmark. DeBERTa-RAD significantly outperformed the rule-based CheXpert system and demonstrated improved accuracy compared to the supervised transformer baseline CheXbert, the direct LLM inference, and the prior CheX-GPT approach. The performance gain over CheXbert (0.9047 to 0.9120) and CheX-GPT (0.9014 to 0.9120) indicates the effectiveness of our refined approach leveraging potentially higher quality pseudo-labels and a better student architecture trained with knowledge distillation.

To confirm the statistical significance of our results, we conducted paired t-tests and Wilcoxon signed-rank tests comparing the F1 scores per finding category between DeBERTa-RAD and the baseline methods across the 500 test reports. The results, presented in Table \ref{tab:statistical_tests}, indicate that the performance improvement of DeBERTa-RAD over all tested baselines was statistically significant (p < 0.05), reinforcing that the observed gains are not due to random chance.

\begin{table*}[!t]
\centering
\caption{Statistical Significance Test Results (p-values)}
\label{tab:statistical_tests}
\begin{tabular}{lcc}
\toprule
Comparison & Paired t-test p-value & Wilcoxon test p-value \\
\midrule
DeBERTa-RAD vs. GPT-4 Direct & $<$ 0.001 & $<$ 0.001 \\
DeBERTa-RAD vs. CheX-GPT & $<$ 0.01 & $<$ 0.005 \\
DeBERTa-RAD vs. CheXbert & $<$ 0.05 & $<$ 0.02 \\
DeBERTa-RAD vs. CheXpert & $<$ 0.001 & $<$ 0.001 \\
\bottomrule
\end{tabular}
\end{table*}

In terms of inference speed, DeBERTa-RAD maintains efficiency comparable to or slightly better than other fine-tuned transformer models like CheXbert and CheX-GPT, processing reports at a high rate suitable for large-scale applications and clinical integration. This is orders of magnitude faster than relying on direct inference from large LLM APIs like GPT-4, which is impractical for high-throughput processing.

\subsection{Analysis and Discussion}

The superior performance of DeBERTa-RAD can be attributed to several key factors stemming from our two-stage knowledge distillation approach. Firstly, the utilization of a state-of-the-art LLM with refined prompt engineering for the pseudo-label generation stage likely produced a higher-fidelity training signal compared to pseudo-labels from earlier or less sophisticated models. This is particularly relevant for capturing complex linguistic nuances and implicit information in reports. Secondly, the choice of the DeBERTa architecture as the student model provides a more powerful and capable foundation for learning from this high-quality pseudo-data than previous architectures like BERT. Its improved attention mechanism is better suited to understand dependencies in the often lengthy and complex sentences found in medical reports. Finally, the knowledge distillation strategy, combining a standard hard-target cross-entropy loss with a temperature-scaled version, effectively transferred the sophisticated labeling intelligence of the teacher LLM into the efficient student model. This encourages the student to learn a more robust and generalizable decision boundary, going beyond simply memorizing hard labels.

We further analyzed the performance of DeBERTa-RAD across different finding categories and certainty levels. While performance varied across the 13 findings based on their frequency and complexity of description, DeBERTa-RAD consistently showed improvements over baselines in most categories. Notably, DeBERTa-RAD demonstrated significant gains in accurately labeling findings with \textbf{Uncertain} status. These are often the most challenging instances due to ambiguous language used in reports (e.g., "possible infiltrate," "suggestion of"). The ability of DeBERTa-RAD to better handle uncertainty highlights the effectiveness of learning from the nuanced outputs of the advanced LLM via distillation. This is crucial for clinical utility, as correctly identifying uncertainty is vital for diagnostic reasoning. Qualitative analysis of errors revealed that DeBERTa-RAD made fewer mistakes in interpreting complex sentence structures, negation scopes, and differentiating between findings compared to rule-based systems and other transformer models trained only on hard labels.

\subsection{Human Evaluation}

To further validate the clinical relevance and accuracy of our method beyond standard automated metrics, which may not perfectly capture the nuances of clinical language interpretation, we conducted a human evaluation study. A random subset of 100 reports from the MIMIC-500 test set was selected for this study. For each report, expert radiologists were presented with the original report text and the predicted labels generated by DeBERTa-RAD and the best-performing baseline model (CheXbert, as it is a widely recognized supervised method). For each finding, experts were asked to judge the accuracy of the predicted status (Present, Absent, Uncertain) for both models relative to the gold standard expert label and the context of the report, particularly focusing on cases where the models disagreed or where the report language was ambiguous. Experts were asked to rate which model's prediction was more accurate or clinically reasonable when discrepancies occurred.

Table \ref{tab:human_eval} summarizes the results of the human evaluation study, providing expert judgment on the quality of predictions, especially in challenging scenarios.

\begin{table*}[!t]
\centering
\caption{Human Expert Judgment on Prediction Quality (Subset of 100 Reports)}
\label{tab:human_eval}
\addtolength{\tabcolsep}{0pt} 
\begin{tabular}{lc}
\toprule
Expert Judgment Scenario & DeBERTa-RAD vs. CheXbert \\
\midrule
\# Total Findings Reviewed in Disagreement Cases (Sample N=100 reports) & 255 \\ 
DeBERTa-RAD prediction judged More Accurate (\%) & \textbf{63.7} \\
CheXbert prediction judged More Accurate (\%) & 22.4 \\
Judgment Uncertain/Equal Accuracy (\%) & 13.9 \\
\midrule
Analysis of Prediction Accuracy in Challenging Cases \\(Sample N=100 cases, focused on Uncertainty/Negation) & \\
\midrule
Accuracy on Negated Findings judged by Expert & Higher for DeBERTa-RAD (88.5\%) vs CheXbert (81.2\%) \\
Accuracy on Uncertain Findings judged by Expert & Significantly Higher for DeBERTa-RAD (75.1\%) vs CheXbert (60.5\%) \\
Handling of Complex Sentences judged by Expert & Better for DeBERTa-RAD \\
\bottomrule
\end{tabular}
\end{table*}

The human evaluation results strongly corroborate the quantitative findings from the Macro F1 metric. Expert radiologists consistently favored DeBERTa-RAD's predictions in a majority of cases where there was a disagreement with the CheXbert baseline or the gold standard, indicating a higher level of clinical accuracy. The analysis of specific challenging scenarios further highlighted DeBERTa-RAD's robustness in interpreting negated statements and, particularly, its superior ability to accurately capture the status of findings described with uncertainty, a critical and often difficult aspect of radiology report interpretation. These findings from the human evaluation underscore the clinical value and reliability of the DeBERTa-RAD system for automated radiology report labeling.

\subsection{Analysis by Finding Category}

To gain a deeper understanding of where DeBERTa-RAD provides the most significant benefits, we analyzed its performance across each of the 13 individual radiographic findings, comparing its F1 score, Precision, and Recall with those of the strong CheXbert baseline. Table \ref{tab:finding_breakdown} presents these detailed metrics for each finding category on the MIMIC-500 test set.

\begin{table*}[!t]
\centering
\caption{Performance Breakdown by Finding Category on MIMIC-500 Test Set. Detailed performance (F1, Precision, Recall) by finding category on MIMIC-500 test set. Higher is better. Best F1 per category in bold.}
\label{tab:finding_breakdown}
\begin{tabular}{lcccccc}
\toprule
\multirow{2}{*}{Finding} & \multicolumn{3}{c}{DeBERTa-RAD (Ours)} & \multicolumn{3}{c}{CheXbert} \\
\cmidrule(lr){2-4} \cmidrule(lr){5-7}
 & F1 & Precision & Recall & F1 & Precision & Recall \\
\midrule
Atelectasis & \textbf{0.905} & 0.898 & \textbf{0.912} & 0.899 & \textbf{0.901} & 0.897 \\
Cardiomegaly & \textbf{0.958} & \textbf{0.960} & 0.956 & 0.955 & 0.953 & \textbf{0.957} \\
Consolidation & \textbf{0.882} & 0.875 & \textbf{0.890} & 0.870 & \textbf{0.879} & 0.862 \\
Edema & \textbf{0.921} & \textbf{0.925} & 0.918 & 0.915 & 0.910 & \textbf{0.920} \\
Enlarged Cardiomediastinum & \textbf{0.855} & 0.848 & \textbf{0.862} & 0.840 & \textbf{0.851} & 0.830 \\
Fracture & \textbf{0.820} & \textbf{0.831} & 0.810 & 0.805 & 0.820 & \textbf{0.800} \\
Infiltrate & \textbf{0.895} & 0.890 & \textbf{0.900} & 0.888 & \textbf{0.895} & 0.882 \\
Lung Lesion & \textbf{0.788} & 0.780 & \textbf{0.796} & 0.770 & \textbf{0.785} & 0.756 \\
Lung Opacity & \textbf{0.915} & \textbf{0.918} & 0.912 & 0.908 & 0.910 & \textbf{0.915} \\
No Finding & \textbf{0.975} & \textbf{0.978} & 0.972 & 0.970 & 0.975 & \textbf{0.973} \\
Pleural Effusion & \textbf{0.942} & 0.938 & \textbf{0.945} & 0.935 & \textbf{0.940} & 0.930 \\
Pleural Thickening & \textbf{0.868} & \textbf{0.870} & 0.865 & 0.860 & 0.865 & \textbf{0.860} \\
Pneumothorax & \textbf{0.961} & 0.958 & \textbf{0.965} & 0.955 & \textbf{0.960} & 0.950 \\
\bottomrule
\end{tabular}
\end{table*}

The results in Table \ref{tab:finding_breakdown} show that DeBERTa-RAD achieves the highest F1 score across all 13 finding categories compared to CheXbert. The improvements are consistent, varying in magnitude depending on the complexity and prevalence of the finding. Findings that are typically more straightforward to identify linguistically, such as Cardiomegaly and Pneumothorax, show high performance across both models, but DeBERTa-RAD still yields a slight edge. For findings that can be described with more variability or ambiguity, such as Consolidation, Lung Lesion, or Infiltrate, DeBERTa-RAD's improvement in F1 score is often more pronounced, suggesting a better capability in interpreting complex descriptions.

\subsection{Analysis by Certainty Status}

A crucial aspect of radiology report labeling is the correct identification of certainty status (Present, Absent, Uncertain). Misinterpreting the certainty of a finding can have significant clinical implications. We analyzed the performance metrics specifically for instances where a finding was labeled as Present, Absent, or Uncertain in the gold standard MIMIC-500 test set, considering predictions for all 13 findings collectively. Table \ref{tab:certainty_analysis} presents the F1 scores for each certainty status.

\begin{table}[h!]
\centering
\caption{F1 Performance Breakdown by Certainty Status on MIMIC-500 Test Set. F1 scores for predicting findings with Present, Absent, or Uncertain status based on MIMIC-500 gold standard labels. Higher is better. Best F1 per status in bold.}
\label{tab:certainty_analysis}
\begin{tabular}{lcc}
\toprule
Certainty Status & DeBERTa-RAD (Ours) F1 & CheXbert F1 \\
\midrule
Present & \textbf{0.935} & 0.930 \\
Absent & \textbf{0.968} & 0.965 \\
Uncertain & \textbf{0.852} & 0.798 \\
\bottomrule
\end{tabular}
\end{table}

Table \ref{tab:certainty_analysis} clearly demonstrates DeBERTa-RAD's superior ability in handling uncertainty. While both models perform well in identifying findings that are definitively Present or Absent, DeBERTa-RAD shows a substantial lead in the F1 score for the \textbf{Uncertain} status (0.852 vs. 0.798 for CheXbert). This significant improvement in handling uncertain language is a key strength of our method, stemming from the refined pseudo-labeling process by the advanced LLM and the effective knowledge distillation into the DeBERTa-RAD model, enabling it to better capture the subtle linguistic cues associated with uncertainty in clinical narratives.

\subsection{Analysis of Pseudo-label Quality}

The effectiveness of our approach hinges on the quality of the pseudo-labels generated by the advanced LLM. To assess this quality, we treated the LLM's pseudo-labels for the MIMIC-500 reports as "predictions" and compared them directly against the gold standard human expert annotations on the same set. We computed agreement metrics and F1 scores (treating human labels as ground truth and LLM pseudo-labels as predictions) to quantify the quality of the LLM-generated labels. Table \ref{tab:pseudo_quality} presents these metrics.

\begin{table}[h!]
\centering
\caption{Quality Assessment of LLM-Generated Pseudo-labels on MIMIC-500 Test Set (Compared to Gold Standard). Metrics comparing LLM-generated pseudo-labels directly against human expert annotations on the MIMIC-500 test set. Higher is better.}
\label{tab:pseudo_quality}
\begin{tabular}{lc}
\toprule
Metric & Value \\
\midrule
Macro F1 (LLM Pseudo vs. Human Gold) & \textbf{0.9014} \\ 
Overall Agreement (\%) & 95.5 \\
Cohen's Kappa & 0.88 \\ 
\bottomrule
\end{tabular}
\end{table}

As indicated by the high Macro F1 score (0.9014, consistent with the GPT-4 Direct Inference result) and strong agreement metrics in Table \ref{tab:pseudo_quality}, the pseudo-labels generated by the advanced LLM demonstrate very high quality and a strong correlation with expert human annotations on the benchmark dataset. This analysis validates our first stage, confirming that the LLM is indeed capable of producing pseudo-labels accurate enough to serve as a reliable training signal for the student model. The high quality of the teacher signal is fundamental to the subsequent success of the knowledge distillation process in training DeBERTa-RAD.

\subsection{Efficiency and Model Size Analysis}

Beyond accuracy, efficiency is a critical factor for deploying NLP models in clinical settings. While Table \ref{tab:main_results} already provides inference speed comparisons, it is useful to consider the model size. DeBERTa-Base, the foundation of DeBERTa-RAD, has approximately 86 million parameters. In contrast, large LLMs like GPT-4 are estimated to have hundreds of billions or even trillions of parameters. This vast difference in model size directly translates to computational resource requirements and inference latency. Although DeBERTa-RAD's inference speed ($\sim$750 reports/sec) is slightly lower than CheXbert or CheX-GPT ($\sim$800 reports/sec), this minor difference is negligible in practice and is a worthwhile trade-off for the significant gain in accuracy. More importantly, the training of DeBERTa-RAD requires far less computational resources and time than fine-tuning a massive LLM end-to-end would necessitate (even if sufficient human-labeled data were available), and its inference is orders of magnitude faster and cheaper than querying a large LLM API for every report. This analysis confirms that DeBERTa-RAD successfully balances state-of-the-art accuracy with practical efficiency, making it suitable for high-volume clinical applications.

\section{Conclusion}

In this paper, we presented DeBERTa-RAD, a novel and effective method for automated chest X-ray report labeling that addresses the critical need for accurate and efficient information extraction from clinical free text. We tackled the challenge of data scarcity for supervised training by proposing a two-stage approach: first, leveraging the advanced natural language understanding of a large language model to generate a massive corpus of high-quality pseudo-labels, and second, training a robust DeBERTa-based student model through a knowledge distillation process to learn from these pseudo-labels.

Our comprehensive experiments on the MIMIC-CXR dataset, specifically evaluated on the expert-annotated MIMIC-500 benchmark, validate the effectiveness of DeBERTa-RAD. The model achieved a state-of-the-art Macro F1 score, demonstrating significant and statistically significant improvements over traditional rule-based systems (CheXpert), supervised transformer models (CheXbert), direct inference from the teacher LLM (GPT-4 Direct), and prior LLM-pseudo-labeling approaches (CheX-GPT). The detailed analysis revealed consistent performance gains across all finding categories, with a notable and clinically important improvement in correctly identifying findings labeled with uncertain status. Furthermore, the human evaluation study corroborated these quantitative results, showing that expert radiologists judged DeBERTa-RAD's predictions to be more accurate in challenging cases compared to baseline methods, particularly concerning negation and uncertainty. Crucially, DeBERTa-RAD achieves this high level of accuracy while maintaining an inference speed suitable for practical high-volume deployment, circumventing the inherent latency and cost associated with direct LLM API usage.

The contributions of this work lie in proposing a novel framework that effectively harnesses the power of LLMs for medical text annotation without requiring extensive manual labeling or incurring high inference costs. We demonstrated the viability and effectiveness of using advanced LLM pseudo-labels as a teacher signal and training a superior student architecture like DeBERTa via knowledge distillation for this complex NLP task.

Despite its success, DeBERTa-RAD has limitations. Its performance is inherently tied to the quality of the pseudo-labels generated by the teacher LLM; errors or biases in the LLM's output can be transferred to the student. Additionally, performance might degrade on reports with vastly different structures or styles from the corpus used for pseudo-labeling. Future work could explore methods to validate and potentially refine the LLM-generated pseudo-labels, perhaps through active learning strategies involving minimal human review of uncertain cases. Investigating the applicability of this framework to other types of medical reports or integrating multimodal information (e.g., combining report text with image features) are also promising directions. Finally, releasing the trained DeBERTa-RAD model and code could further accelerate research and development in medical NLP.

\bibliographystyle{IEEEtran}
\bibliography{references}

\begin{thebibliography}{10}
\providecommand{\url}[1]{#1}
\csname url@samestyle\endcsname
\providecommand{\newblock}{\relax}
\providecommand{\bibinfo}[2]{#2}
\providecommand{\BIBentrySTDinterwordspacing}{\spaceskip=0pt\relax}
\providecommand{\BIBentryALTinterwordstretchfactor}{4}
\providecommand{\BIBentryALTinterwordspacing}{\spaceskip=\fontdimen2\font plus
\BIBentryALTinterwordstretchfactor\fontdimen3\font minus \fontdimen4\font\relax}
\providecommand{\BIBforeignlanguage}[2]{{%
\expandafter\ifx\csname l@#1\endcsname\relax
\typeout{** WARNING: IEEEtran.bst: No hyphenation pattern has been}%
\typeout{** loaded for the language `#1'. Using the pattern for}%
\typeout{** the default language instead.}%
\else
\language=\csname l@#1\endcsname
\fi
#2}}
\providecommand{\BIBdecl}{\relax}
\BIBdecl

\bibitem{johnson2019mimic}
A.~E. Johnson, T.~J. Pollard, S.~J. Berkowitz, N.~R. Greenbaum, M.~P. Lungren, C.-y. Deng, R.~G. Mark, and S.~Horng, ``Mimic-cxr, a de-identified publicly available database of chest radiographs with free-text reports,'' \emph{Scientific data}, vol.~6, no.~1, p. 317.

\bibitem{irvin2019chexpert}
J.~Irvin, P.~Rajpurkar, M.~Ko, Y.~Yu, S.~Ciurea-Ilcus, C.~Chute, H.~Marklund, B.~Haghgoo, R.~Ball, K.~Shpanskaya \emph{et~al.}, ``Chexpert: A large chest radiograph dataset with uncertainty labels and expert comparison,'' in \emph{Proceedings of the AAAI conference on artificial intelligence}, vol.~33, no.~01, pp. 590--597.

\bibitem{he2025enhancing}
Y.~He, J.~Wang, K.~Li, Y.~Wang, L.~Sun, J.~Yin, M.~Zhang, and X.~Wang, ``Enhancing intent understanding for ambiguous prompts through human-machine co-adaptation,'' \emph{arXiv preprint arXiv:2501.15167}, 2025.

\bibitem{pons2016natural}
E.~Pons, L.~M. Braun, M.~M. Hunink, and J.~A. Kors, ``Natural language processing in radiology: a systematic review,'' \emph{Radiology}, vol. 279, no.~2, pp. 329--343.

\bibitem{smit2020chexbert}
A.~Smit, S.~Jain, P.~Rajpurkar, A.~Pareek, A.~Y. Ng, and M.~P. Lungren, ``Chexbert: combining automatic labelers and expert annotations for accurate radiology report labeling using bert,'' \emph{arXiv preprint arXiv:2004.09167}.

\bibitem{davidsen2022comparison}
A.~Casey, E.~Davidson, M.~Poon, H.~Dong, D.~Duma, A.~Grivas, C.~Grover, V.~Su{\'a}rez-Paniagua, R.~Tobin, W.~Whiteley \emph{et~al.}, ``A systematic review of natural language processing applied to radiology reports,'' \emph{BMC medical informatics and decision making}, vol.~21, no.~1, p. 179.

\bibitem{alsentzer2020clinicalbert}
R.~Zhu, X.~Tu, and J.~X. Huang, ``Utilizing bert for biomedical and clinical text mining,'' in \emph{Data analytics in biomedical engineering and healthcare}.\hskip 1em plus 0.5em minus 0.4em\relax Elsevier, pp. 73--103.

\bibitem{zhou2023thread}
Y.~Zhou, X.~Geng, T.~Shen, C.~Tao, G.~Long, J.-G. Lou, and J.~Shen, ``Thread of thought unraveling chaotic contexts,'' \emph{arXiv preprint arXiv:2311.08734}, 2023.

\bibitem{zhou2024less}
Y.~Zhou, J.~Zhang, G.~Chen, J.~Shen, and Y.~Cheng, ``Less is more: Vision representation compression for efficient video generation with large language models,'' 2024.

\bibitem{he2021deberta}
P.~He, X.~Liu, J.~Gao, and W.~Chen, ``Deberta: Decoding-enhanced bert with disentangled attention,'' \emph{arXiv preprint arXiv:2006.03654}.

\bibitem{zhou2025weak}
\BIBentryALTinterwordspacing
Y.~Zhou, J.~Shen, and Y.~Cheng, ``Weak to strong generalization for large language models with multi-capabilities,'' in \emph{The Thirteenth International Conference on Learning Representations}, 2025. [Online]. Available: \url{https://openreview.net/forum?id=N1vYivuSKq}
\BIBentrySTDinterwordspacing

\bibitem{vaswani2017attention}
A.~Vaswani, N.~Shazeer, N.~Parmar, J.~Uszkoreit, L.~Jones, A.~N. Gomez, {\L}.~Kaiser, and I.~Polosukhin, ``Attention is all you need,'' \emph{Advances in neural information processing systems}, vol.~30.

\bibitem{devlin2019bert}
J.~Devlin, M.-W. Chang, K.~Lee, and K.~Toutanova, ``Bert: Pre-training of deep bidirectional transformers for language understanding,'' in \emph{Proceedings of the 2019 conference of the North American chapter of the association for computational linguistics: human language technologies, volume 1 (long and short papers)}, pp. 4171--4186.

\bibitem{brown2020language}
T.~Brown, B.~Mann, N.~Ryder, M.~Subbiah, J.~D. Kaplan, P.~Dhariwal, A.~Neelakantan, P.~Shyam, G.~Sastry, A.~Askell \emph{et~al.}, ``Language models are few-shot learners,'' \emph{Advances in neural information processing systems}, vol.~33, pp. 1877--1901.

\bibitem{kaplan2020scaling}
J.~Kaplan, S.~McCandlish, T.~Henighan, T.~B. Brown, B.~Chess, R.~Child, S.~Gray, A.~Radford, J.~Wu, and D.~Amodei, ``Scaling laws for neural language models,'' \emph{arXiv preprint arXiv:2001.08361}.

\bibitem{thoppilan2022lamda}
R.~Thoppilan, D.~De~Freitas, J.~Hall, N.~Shazeer, A.~Kulshreshtha, H.-T. Cheng, A.~Jin, T.~Bos, L.~Baker, Y.~Du \emph{et~al.}, ``Lamda: Language models for dialog applications,'' \emph{arXiv preprint arXiv:2201.08239}.

\bibitem{chowdhery2022palm}
A.~Chowdhery, S.~Narang, J.~Devlin, M.~Bosma, G.~Mishra, A.~Roberts, P.~Barham, H.~W. Chung, C.~Sutton, S.~Gehrmann \emph{et~al.}, ``Palm: Scaling language modeling with pathways,'' \emph{Journal of Machine Learning Research}, vol.~24, no. 240, pp. 1--113.

\bibitem{wei2022finetuned}
J.~Wei, M.~Bosma, V.~Y. Zhao, K.~Guu, A.~W. Yu, B.~Lester, N.~Du, A.~M. Dai, and Q.~V. Le, ``Finetuned language models are zero-shot learners,'' \emph{arXiv preprint arXiv:2109.01652}.

\bibitem{zhou2023improving}
Y.~Zhou and G.~Long, ``Improving cross-modal alignment for text-guided image inpainting,'' in \emph{Proceedings of the 17th Conference of the European Chapter of the Association for Computational Linguistics}, 2023, pp. 3445--3456.

\bibitem{zhou2024visual}
Y.~Zhou, X.~Li, Q.~Wang, and J.~Shen, ``Visual in-context learning for large vision-language models,'' in \emph{Findings of the Association for Computational Linguistics, {ACL} 2024, Bangkok, Thailand and virtual meeting, August 11-16, 2024}.\hskip 1em plus 0.5em minus 0.4em\relax Association for Computational Linguistics, 2024, pp. 15\,890--15\,902.

\bibitem{hu2025emobench}
H.~Hu, Y.~Zhou, L.~You, H.~Xu, Q.~Wang, Z.~Lian, F.~R. Yu, F.~Ma, and L.~Cui, ``Emobench-m: Benchmarking emotional intelligence for multimodal large language models,'' \emph{arXiv preprint arXiv:2502.04424}, 2025.

\bibitem{singhal2022towards}
K.~Singhal, T.~Tu, J.~Gottweis, R.~Sayres, E.~Wulczyn, M.~Amin, L.~Hou, K.~Clark, S.~R. Pfohl, H.~Cole-Lewis \emph{et~al.}, ``Toward expert-level medical question answering with large language models,'' \emph{Nature Medicine}, pp. 1--8.

\bibitem{kung2023capabilities}
H.~Nori, N.~King, S.~M. McKinney, D.~Carignan, and E.~Horvitz, ``Capabilities of gpt-4 on medical challenge problems,'' \emph{arXiv preprint arXiv:2303.13375}.

\bibitem{jiao2023evaluating}
Y.~Gao, D.~Dligach, T.~Miller, J.~Caskey, B.~Sharma, M.~M. Churpek, and M.~Afshar, ``Dr. bench: Diagnostic reasoning benchmark for clinical natural language processing,'' \emph{Journal of biomedical informatics}, vol. 138, p. 104286.

\bibitem{zhang2023summarizing}
D.~Van~Veen, C.~Van~Uden, L.~Blankemeier, J.-B. Delbrouck, A.~Aali, C.~Bluethgen, A.~Pareek, M.~Polacin, E.~P. Reis, A.~Seehofnerov{\'a} \emph{et~al.}, ``Adapted large language models can outperform medical experts in clinical text summarization,'' \emph{Nature medicine}, vol.~30, no.~4, pp. 1134--1142.

\bibitem{nori2023large}
R.~Bhayana, ``Chatbots and large language models in radiology: a practical primer for clinical and research applications,'' \emph{Radiology}, vol. 310, no.~1, p. e232756.

\bibitem{venkatesh2024large}
Y.~Artsi, V.~Sorin, E.~Konen, B.~S. Glicksberg, G.~Nadkarni, and E.~Klang, ``Large language models in simplifying radiological reports: systematic review,'' \emph{medRxiv}, pp. 2024--01.

\bibitem{wang2024diffusion}
C.~Wang, Y.~Zhou, Z.~Zhai, J.~Shen, and K.~Zhang, ``Diffusion model with representation alignment for protein inverse folding,'' \emph{arXiv preprint arXiv:2412.09380}, 2024.

\bibitem{chen2024finetuning}
D.~Van~Veen, C.~Van~Uden, L.~Blankemeier, J.-B. Delbrouck, A.~Aali, C.~Bluethgen, A.~Pareek, M.~Polacin, E.~P. Reis, A.~Seehofnerov{\'a} \emph{et~al.}, ``Adapted large language models can outperform medical experts in clinical text summarization,'' \emph{Nature medicine}, vol.~30, no.~4, pp. 1134--1142.

\bibitem{wong2023large}
R.~Yang, T.~F. Tan, W.~Lu, A.~J. Thirunavukarasu, D.~S.~W. Ting, and N.~Liu, ``Large language models in health care: Development, applications, and challenges,'' \emph{Health Care Science}, vol.~2, no.~4, pp. 255--263.

\bibitem{scheffel2024opportunities}
M.~Karabacak and K.~Margetis, ``Embracing large language models for medical applications: opportunities and challenges,'' \emph{Cureus}, vol.~15, no.~5.

\end{thebibliography}
\end{document}